\documentclass[runningheads]{llncs}
\usepackage{graphicx}
\usepackage{amsmath}
\usepackage{amssymb}
\usepackage{graphicx}
\usepackage{booktabs}
\usepackage{arydshln}
\usepackage{multirow}
\usepackage{tablefootnote}
\usepackage{tabularx} 

\begin{document}
\title{A Task-oriented Dialog Model with Task-progressive and Policy-aware Pre-training}
\titlerunning{A task-progressive PCM with Two Policy-aware Pre-training Tasks}

\author{Lucen Zhong\inst{1} \and
Hengtong Lu\inst{1} \and
Caixia Yuan\inst{1} \and
Xiaojie Wang\inst{1} \and
Jiashen Sun\inst{2} \and
Ke Zeng\inst{2} \and
Guanglu Wan\inst{2}
}

\authorrunning{L. Zhong et al.}

\institute{Center of Intelligence Science and Technology, Beijing University of Posts and Telecommunications, China \\
\email{\{zhonglucen, luhengtong, yuancx, xjwang\}@bupt.edu.cn} \\
\and 
Meituan, China \\
\email{\{sunjiashen, zengke02, wanguanglu\}@meituan.com}
}

\maketitle              
\begin{abstract}
Pre-trained conversation models (PCMs) have achieved promising progress in recent years. However, existing PCMs for Task-oriented dialog (TOD)  are insufficient for capturing the sequential nature of the TOD-related tasks, as well as for learning dialog policy information. To alleviate these problems, this paper proposes a task-progressive PCM with two policy-aware pre-training tasks. The model is pre-trained through three stages where TOD-related tasks are progressively employed according to the task logic of the TOD system. A global policy consistency task is designed to capture the multi-turn dialog policy sequential relation, and an act-based contrastive learning task is designed to capture similarities among samples with the same dialog policy. Our model achieves better results on both MultiWOZ and In-Car end-to-end dialog modeling benchmarks with only 18\% parameters and 25\% pre-training data compared to the previous state-of-the-art PCM, GALAXY. We make our code and data publicly available. \footnote{https://github.com/lucenzhong/TPLD}

\end{abstract}
\section{Introduction}
Task-oriented dialog (TOD) system aims at helping users complete specific tasks through multi-turn interactions. Compared with open domain dialog agents, a TOD system generates more controllable replies by implementing three sub-tasks: 1) Dialog State Tracking (DST) extracts the belief state; 2) Dialog Policy Learning (POL) decides which acts should be taken based on the belief state; 3) Natural Language Generation (NLG) converts acts into natural language utterances. A large amount of work has been done for each sub-task \cite{tian2021amendable,takanobu2020multi,peng-etal-2020-shot} separately, as well as joint models for them \cite{madotto-etal-2018-mem2seq,lei-etal-2018-sequicity}. 

Pre-trained Conversation Models (PCMs) \cite{peng-etal-2021-soloist,he2022galaxy,he2022space,he2022unified} are Pre-trained Language Models (PLMs) further pre-trained on dialog data. Although previous work on PCMs for TOD has made big progress, the following issues are still not well-addressed: 
1) When TOD-related sub-tasks are used as pre-training tasks for PCMs, they are always employed simultaneously in a multi-task way. However, DST, POL and NLG are essentially sequential tasks in the TOD system. Managing sequential tasks in a multi-task way cannot capture the sequential nature of these tasks and it is difficult to better learn the subsequent task due to insufficient learning of the previous task. 
2) Existing work only optimizes the policy for each dialog turn \cite{he2022unified}. However, TOD is essentially a multi-turn sequential decision-making process, so it is more critical to build pre-training tasks that learn to optimize dialog policy over the whole dialog. In addition, existing work only models the policy differences between samples in the same batch, ignoring the similarities among samples with the same policy in data sets.

\begin{figure}[] 
\centering 
\includegraphics[width=0.95\textwidth]{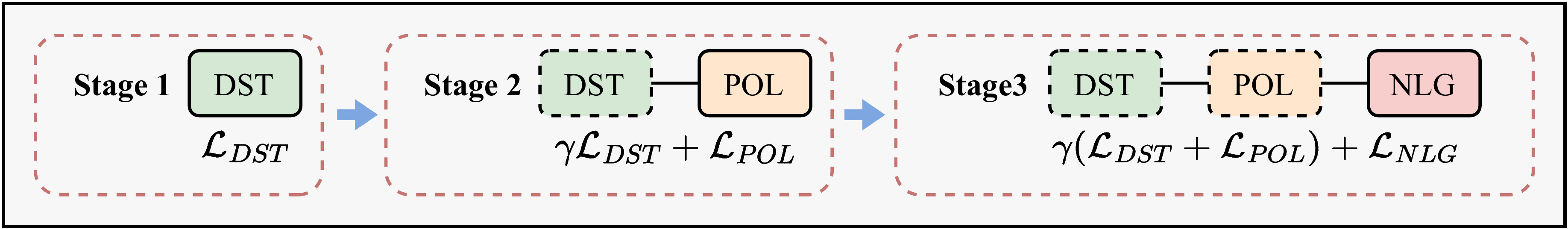} 
\caption{The TPLD multi-stage pre-training framework.}
\label{task-progressive_with_decay}
\vspace{-0.3cm}
\end{figure}

To address the above problems, this paper first proposes a \textbf{T}ask-\textbf{P}rogressive with \textbf{L}oss \textbf{D}ecaying (TPLD) multi-stage pre-training framework for training TOD PCMs. As shown in Figure \ref{task-progressive_with_decay}, the framework includes three stages of pre-training. DST, POL, and NLG tasks are progressively employed in different stages according to the task logic of the TOD system. Since DST, POL, and NLG tasks for PCMs are heterogeneous tasks, the latter task may completely offset the tasks in the previous stages. Therefore, tasks employed in previous stages are assigned a decayed loss in the current stage. The decayed loss is used to leverage tasks from the previous stage so that current tasks can not compeletely offset the previous task. At the same time, we propose two policy-aware pre-training tasks to enhance policy learning. A global policy consistency task, which minimizes the ${L}_{2}$ distance between the policy prior and policy posterior both at the turn-level and the session-level, is proposed to model both the single and multiple turn policy. We also propose an act-based contrastive learning task by introducing out-of-batch positive samples to learn the similarities between dialogs with the same policy and the differences between dialogs with different policies simultaneously.

T5-small \cite{raffel2020exploring} is employed as the backbone model. Experimental results show that our model outperforms previous state-of-the-art PCM on both MultiWOZ and In-Car end-to-end dialog modeling benchmarks. 
In summary, the main contributions of the paper are as follows:
\begin{enumerate}
    \item We propose a task-progressive pre-training framework for TOD PCMs, which leverages sequential nature between the different pre-training tasks.
    \item We propose two novel and effective policy-aware pre-training tasks for dialog policy modeling. To the best of our knowledge, it is the first session-level dialog policy pre-training task.
    \item Our model achieves better results on two popular end-to-end dialog modeling benchmarks with fewer parameters and less pre-training data compared with previous strong PCMs.
\end{enumerate}

\section{Related work}
\textbf{Pre-trained Language Models for TOD.} Pre-trained Language Models (PLMs) trained on large general text corpora \cite{radford2019language,raffel2020exploring}, have been widely applied to dialog systems \cite{yang2021ubar,sun2022mars}. UBAR \cite{yang2021ubar} evaluates the task-oriented dialog system in a more realistic setting, where its dialog context has access to user utterances and all generated content. Mars \cite{sun2022mars} proposes two contrastive learning strategies to model the dialog context and belief/action state.  Since the intrinsic linguistic patterns differ between dialog and normal text, PLMs-based TOD models achieve limited progress.
\\[3pt]
\noindent\textbf{Pre-trained Conversation Models.} In order to bridge the gap caused by pre-training data, some studies \cite{henderson-etal-2020-convert,adiwardana2020towards} further pre-trained the PLMs on dialog corpora to build pre-trained conversation models(PCMs). Many PCMs are trained on open-domain dialog data for response generation, and here we concentrate on PCMs for TOD. SC-GPT \cite{peng-etal-2020-shot} first exploited pre-train PLMs for the NLG module in TOD systems. TOD-BERT \cite{wu-etal-2020-tod} and SPACE-2 \cite{he2022space} trained a dialog understanding model that can accomplish tasks like intent recognition and state tracking. SOLOIST \cite{peng-etal-2021-soloist} pre-trained a task-grounded response generation model, which can generate dialog responses grounded in user goals and real-world knowledge for task completion. PPTOD \cite{su-etal-2022-multi} introduced a multi-task pre-training strategy that augments the model’s ability with heterogeneous dialog corpora. GALAXY \cite{he2022galaxy} proposed to learn turn-level dialog policy from limited labeled dialogs and large-scale unlabeled dialog corpora. SPACE-3 \cite{he2022unified} combined GALAXY and SPACE-2 to propose a unified model for dialog understanding and generation. OPAL \cite{chen2022opal} leveraged external tools to generate TOD-like data to bridge the gap between pre-training and fine-tuning. Existing work did not explore pre-training methods other than multi-task learning and only learned turn-level policy. 

\section{Method}

\begin{figure*}[] 
\centering 
\includegraphics[width=.95 \textwidth]{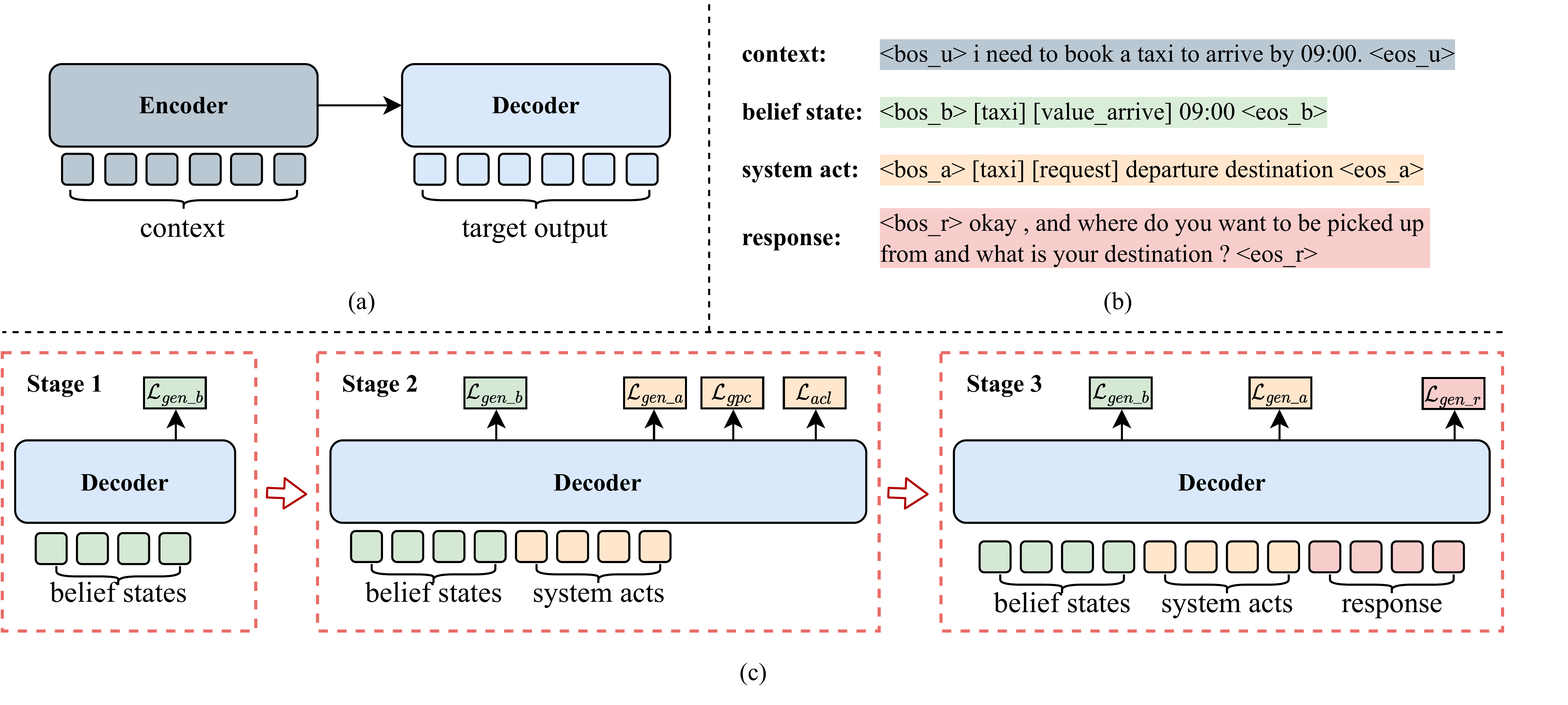}
\vspace{-0.1cm}
\caption{Overview of our proposed method. (a) is the general T5 architecture we use as the backbone. (b) is the data format in the model. (c) is the TPLD multi-stage pre-training framework, where different color for different pre-training tasks.}
\label{model_architecture}
\vspace{-0.3cm}
\end{figure*}

In this section, we first introduce the \textbf{T}ask-\textbf{P}rogressive with \textbf{L}oss \textbf{D}ecaying (TPLD) multi-stage pre-training framework for training TOD PCMs. Then we describe two policy-aware pre-training tasks. Figure \ref{model_architecture} gives some overview information of our method.

\subsection{TPLD Multi-stage Pre-training Framework}
The pre-training process of the model is divided into three stages. DST, POL, and NLG tasks are introduced stage by stage, considering the sequential nature of these tasks in the TOD system. 

Specifically, only a generative DST task that generates the belief states of dialogs is employed in the first stage. Then some POL tasks (including one generative POL task that generates the system acts) are joined in the second stage. We remain the generative DST task together with newly joined POL tasks in the second stage to prevent the model from forgetting the DST task learned in the first stage. At the same time, to make the model focus more on newly joined POL tasks, we multiply the loss function of the generative DST task by a decaying coefficient $\gamma \in [0, 1]$ to weaken its impact. Finally, the NLG task is joined in the third stage. The same decaying coefficient applies to loss functions of both the generative DST and POL tasks. NLG is naturally a generative task that generates the system response. 

We give a formal description of the process as follows: We first define a general form for all three generative pre-training tasks and their loss functions. Then, we introduce the loss function stage by stage in the following subsections. 

A training sample is denoted as in equation (\ref{sample}):
\begin{equation}
\label{sample}
    d = \left ( c,y  \right )
\end{equation}
where $c$ denotes the input dialog context, which is the concatenation of all previous utterances in the dialog. $y$ is the target output text. It is different from different tasks. e.g., it is the belief state in the generative DST task, and the system act in the generative POL task. 

Given the training sample $d$, the generation loss $\mathcal{L}_{gen} $ is as in equation (\ref{generation}):
\begin{equation}
\label{generation}
    \mathcal{L}_{gen} = \sum_{i=1}^{\left |  y\right | } \log_{}{P_{\Theta } \left ( y_{i} | y_{<i},c  \right ) } 
\end{equation}
where $\Theta$ is the model parameter and $y_{<i}$ indicates all tokens before $i$.

\subsubsection{Stage 1: DST Pre-training}
The first stage includes only one generative DST task. The output $y$ is the belief state, and the loss is denoted as $\mathcal{L}_{gen\_b}$. The pre-training objective function for the first stage is as in equation (\ref{stage_1}):
\begin{equation}
\label{stage_1}
    \mathcal{L}_{stage\_1}= \mathcal{L}_{gen\_b}
\end{equation}

\subsubsection{Stage 2: DST+POL Pre-training}
Three POL tasks are joined in the second stage. One of the POL tasks is the system act generation task, where the output $y$ is the system act. The loss function of the task is denoted as $\mathcal{L}_{gen\_a}$. The other two POL tasks, the global policy consistency task with loss function of $\mathcal{L}_{gpc}$ and the act-based contrastive learning task with loss function of $\mathcal{L}_{acl}$, are described in Section \ref{subsection: Policy-aware Pre-training Tasks} in details. The final training objective function for the second stage is as in equation (\ref{stage_2}):
\begin{equation}
\label{stage_2}
\begin{split}
    \mathcal{L}_{stage\_2} &= \gamma \mathcal{L}_{gen\_b} + ( \mathcal{L}_{gen\_a} + \alpha \mathcal{L}_{gpc} + \beta \mathcal{L}_{acl})
\end{split}
\end{equation}
where $\gamma \in [0, 1]$ is the decaying coefficient leveraging DST and POL tasks, $\alpha\in [0, 1]$ and $\beta\in [0, 1]$ are used to leverage different POL tasks. 

\subsubsection{Stage 3: DST+POL+NLG Pre-training}
The NLG task is joined in the third stage. The output $y$ for the NLG task is the delexicalized system response, and the loss function is denoted as $\mathcal{L}_{gen\_r}$. The training objective function for the third stage is as in equation (\ref{stage_3}):
\begin{equation}
\label{stage_3}
    \mathcal{L}_{stage\_3}= \gamma \left ( \mathcal{L}_{gen\_b} + \mathcal{L}_{gen\_a} \right ) + \mathcal{L}_{gen\_r}
\end{equation}
where $\gamma$ is the same decaying coefficient as that in equation (\ref{stage_2}). Please note that the $\gamma$ only act on the generative task.     

\subsection{Policy-aware Pre-training Tasks}
\label{subsection: Policy-aware Pre-training Tasks}

\subsubsection{Global Policy Consistency Task.} 
\begin{figure}[] 
\centering 
\includegraphics[width=0.9\textwidth]{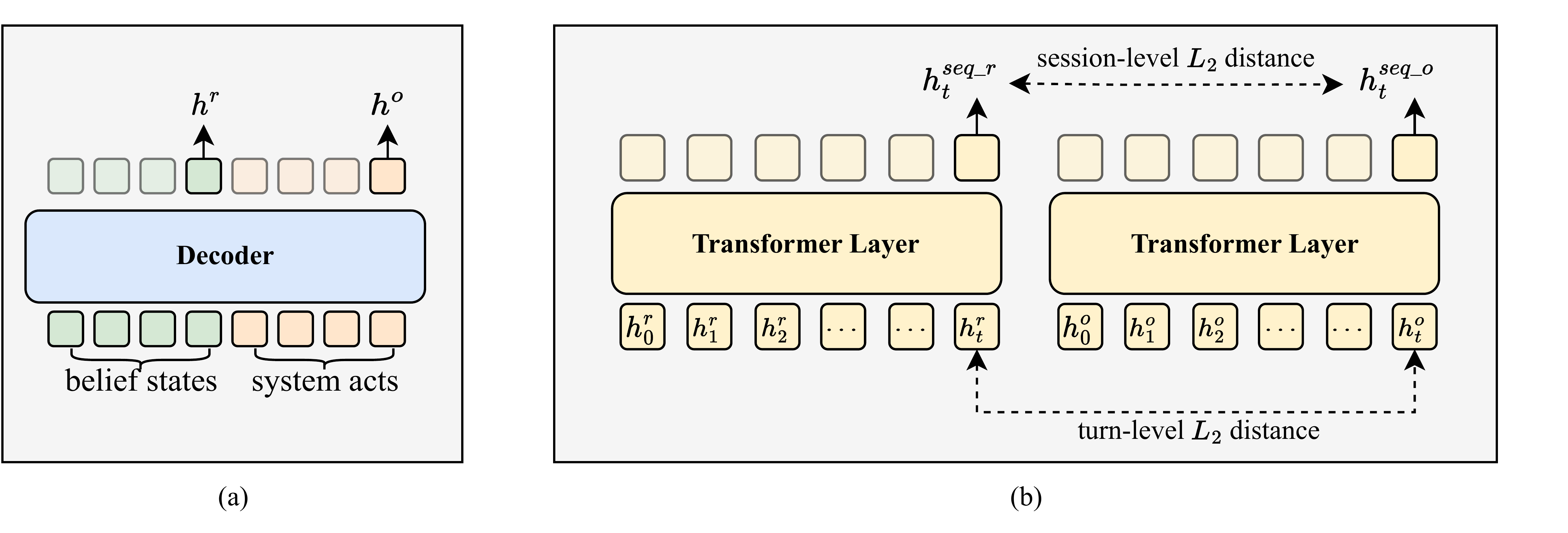} 
\vspace{-0.3cm}
\caption{(a) is the illustration of the policy prior and posterior. (b) is the illustration of global policy consistency task.}
\label{l2_loss}
\vspace{-0.2cm}
\end{figure}

As shown in Figure \ref{l2_loss}(a), we denote the output of the last token in belief states as the policy prior $h^{r}$, and the output of the last token in system acts as the policy posterior $h^{o}$. The dialog policy is unknown in the former and known in the later. Following He et al. \cite{he2022unified}, the turn-level consistency task is to minimizing the $L_{2}$ distance between the representation of the prior and the posterior:
\begin{equation}
    \mathcal{L}_{turn} = \left \| h^{r}_{t} - h^{o}_{t} \right \| _{2}^{2} 
\end{equation}

We further define the session-level loss function for training the global policy consistency task as shown in Figure \ref{l2_loss}(b). Let the prior and the posterior of the policy vector at turn $t$ be $h^{r}_{t}$ and $h^{o}_{t}$, respectively. We can have the policy prior sequence $\left \{ h_{0}^{r},  h_{1}^{r}, \dots , h_{t}^{r} \right \} $ and the policy posterior sequence $\left \{ h_{0}^{o},  h_{1}^{o}, \dots , h_{t}^{o} \right \} $ in hand with the dialog steps forward. A single transformer layer is used to transform the policy sequence into a policy sequence representation for both prior and the posterior policy as shown in equation (7) and (8):
\begin{equation}
    h_{t}^{seq\_r} = Transformer \left ( h_{0}^{r},  h_{1}^{r}, \dots , h_{t}^{r} \right )
\end{equation}
\begin{equation}
    h_{t}^{seq\_o} = Transformer \left ( h_{0}^{o},  h_{1}^{o}, \dots , h_{t}^{o} \right )
\end{equation}

The session-level consistency task is to minimizing the $L_{2}$ distance between the representation of the prior sequence and the posterior sequence:
\begin{equation}
    \mathcal{L}_{session} = \left \| h^{seq\_r}_{t} - h^{seq\_o}_{t} \right \| _{2}^{2}
\end{equation}
The training objective for the global policy consistency task is the sum of turn-level and session-level objectives, as shown in equation (\ref{gpc}):
\begin{equation}
\label{gpc}
    \mathcal{L}_{gpc} = \mathcal{L}_{turn} + \mathcal{L}_{session}
\end{equation}
The turn-level objective models the single-turn dialog policy, while the session-level objective models the multi-turn dialog policy.

\subsubsection{Act-based Contrastive Learning Task.} 
The act-based contrastive learning task aims to introduce out-of-batch positive samples. We treat all samples in the same batch as negative ones and select samples with the same dialog policy from the whole dataset as positive ones. The batch size is denoted as $N$, given a batch of training samples $D = \left \{ d_{1}, d_{2},\cdots , d_{N}    \right \} $, we select $M$ positive samples for each sample $d_{i}$ in this batch and get a new batch with $(M+1)N$ size. Let $I = \left \{ 1, \dots  ,\left ( M+1 \right )N \right \}$ be the index set of the new batch. The act-based contrastive learning loss adopts the policy prior vector $h^{r}$ as the sample vector $h$ and the learning objective is defined as in equation (\ref{acl}):
\begin{equation}
\label{acl}
\small{
    \mathcal{L}_{a c l}= - \sum_{i \in I} \sum_{j \in P_{i}} \log \frac{\exp \left(\sigma\left(h_{i}\right) \cdot \sigma\left(h_{j}\right) / \tau\right)}{  \sum_{l \in I, l \neq i} \exp \left(\sigma\left(h_{i}\right) \cdot \sigma\left(h_{l}\right) / \tau\right)}
}
\end{equation}
where $P_{i}$ is a list of size $M$ which denotes all the positive samples of sample $i$ in the current batch.  $\tau$ is a temperature hyper-parameter. The act-based contrastive learning task can learn the similarities between samples with the same dialog policy and the differences between samples with different dialog policies simultaneously.

\subsection{Fine-tuning and Inference}
In the fine-tuning stage, we focus on the end-to-end dialog modeling task in the TOD system. We only use the generation task during fine-tuning, and the target output $y$ is the concatenation of the belief state, system act, and delexicalized response. The training objective function for fine-tuning is as in equation (\ref{finetune}):
\begin{equation}
\label{finetune}
    \mathcal{L}_{fine\_tune}= \mathcal{L}_{gen\_b} + \mathcal{L}_{gen\_a} + \mathcal{L}_{gen\_r}
\end{equation}
Note that $\mathcal{L}_{gen\_b}$ and $\mathcal{L}_{gen\_a}$ are optional since some datasets do not have corresponding semantic labels.

In the inference stage, following Yang et al. \cite{yang2021ubar}, we use generated system response instead of oracle system response in the context to generate the current system response.

\section{Experiment Settings}

\subsection{Pre-training datasets}.  

Five existing high-quality labeled TOD datasets are used for pre-training our model, including MultiWOZ \cite{budzianowski-etal-2018-multiwoz}, KVRET \cite{eric-etal-2017-key}, MSRE2E \cite{li2018microsoft}, Frames \cite{el-asri-etal-2017-frames}, and CamRest676 \cite{wen-etal-2017-network}. In order to reduce the label discrepancy between different datasets, we follow the unified DA taxonomy \cite{he2022galaxy} to unify the dialog act annotations and use the semantic meaning of slot to unify the slot name annotations. 
Compared with other PCMs, our model uses the least data, with only 25\% of the pre-training data compared to GALAXY.

\subsection{Evaluation Tasks and Metrics}
We test our model on two popular TOD benchmarks: Stanford In-Car Assistant (In-Car) dataset \cite{eric-etal-2017-key} and the MultiWOZ dataset \cite{eric2019multiwoz}. Following previous work \cite{yang2021ubar,peng-etal-2021-soloist}, the model generates delexicalized responses. \texttt{BLEU} score \cite{papineni-etal-2002-bleu} is used to measure the response quality. For MultiWOZ, \texttt{Inform} and \texttt{Success} \cite{budzianowski-etal-2018-multiwoz} are also reported to measure the dialog completion. A \texttt{Combined score} \cite{mehri-etal-2019-structured} is computed by (\texttt{Inform} $+$ \texttt{Success}) $\times0.5 + $ \texttt{BLEU} as an overall quality measure. In order to make a fair comparison with previous work, we adopt the standard evaluation script \cite{nekvinda-dusek-2021-shades} for the evaluation of the MultiWOZ dataset. Similarly, we calculate \texttt{Match}, \texttt{SuccF1} \cite{lei-etal-2018-sequicity}, and the \texttt{Combined score} via (\texttt{Match} $+$ \texttt{SuccF1}) $\times0.5 + $ \texttt{BLEU} for the In-car dataset.

\subsection{Baselines}
We compare our model with the state-of-the-art PCMs for TOD: 
1) \textbf{SOLOIST} \cite{peng-etal-2021-soloist} is a GPT-based model that has been further pre-trained on two TOD datasets; 2) \textbf{PPTOD} \cite{su-etal-2022-multi} is a T5-based model that has been continually pre-trained on eleven heterogeneous annotated TOD corpora; 3) \textbf{GALAXY} \cite{he2022galaxy} is a UniLM-based dialog model that explicitly learns dialog policy from labeled dialogs and large-scale unlabeled dialog corpora via semi-supervised learning; 4) \textbf{SPACE-3} \cite{he2022unified} is a unified semi-supervised pre-trained conversation model learning from large-scale dialog corpora.

\subsection{Implementation Details}
We employ t5-small as the backbone. In the pre-training stage, our model is trained for about 12 hours on one A100 GPU. We use the Adam optimizer \cite{kingma2014adam} with a learning rate of 5e-4 and a batch size of 16 for 15 epochs at each stage. For the hyper-parameters of loss coefficients, we set $\alpha = 0.1 $, $\beta = 1$, and $\gamma = 0.1$, respectively. For hyper-parameters of the act-based contrastive learning task, we set $M = 2$ and $\tau = 1.0$. We removed the validation and testing set of MultiWOZ and In-car during pre-training to avoid a data breach. In the fine-tuning stage, for the MultiWOZ dataset, the learning rate is 5e-4, and the batch size is 16. For the In-Car dataset, the learning rate is 1e-3, and the batch size is 32. We fine-tune the pre-trained model on each dataset for 10 epochs and select the best model based on the validation results. Our implementation is based on the Huggingface Library \cite{wolf2019huggingface}.

\section{Experiment Results}

\subsection{Result Comparisons}
As shown in Table \ref{tab: e2e}, compared with other PCMs, our model achieves new state-of-the-art combined scores on both datasets, outperforms the previous SOTA by 2.0 and 1.2 points on MultiWOZ and In-Car respectively. 
In particular, it is worth noticing that our model surpasses GALAXY, the current best dialog policy learning PCM with explicit policy injection, by 1.9 \texttt{Success} rate and 0.3 \texttt{SuccF1} rate for MultiWOZ and In-Car, respectively. The higher dialog success rates of our model demonstrate that our model can learn better dialog policy than other models to facilitate the completion of dialog tasks.

\begin{table}[htb]
    \centering
    \caption{The Performances on MultiWOZ and In-Car dataset\tablefootnote{We do not compare with SPACE-3 on MultiWOZ because it did not report results on the standard MultiWOZ evaluation script.}}
    \label{tab: e2e}
    \scalebox{0.8}{
    \begin{tabular}{l|cccc|cccc}
    \toprule
    \multirow{2}*{Model} & \multicolumn{4}{c|}{MultiWOZ} & \multicolumn{4}{c}{In-Car} \\ 
    & \texttt{Inform} & \texttt{Success} & \texttt{BLEU} & \texttt{Comb} & \texttt{Match} & \texttt{SuccF1} & \texttt{BLEU} & \texttt{Comb} \\ 
    \midrule
    SOLOIST & 82.3 & 72.4 & 13.6 & 90.9 & - & - & - & - \\
    PPTOD & 83.1 & 72.7 & 18.2 & 96.1 & - & - & - & 106.0 \\
    GALAXY & 85.4 & 75.7 & \textbf{19.6} & 100.2 & 85.3 & 83.6 & 23.0 & 107.5 \\
    SPACE-3 & - & - & - & - & 85.2 & 83.1 & 22.9 & 107.1 \\
    ours & \textbf{89.5} & \textbf{77.6} & 18.7 & \textbf{102.2} & \textbf{86.2} & \textbf{83.9} & \textbf{23.6} & \textbf{108.7} \\ 
    \bottomrule
    \end{tabular}
    }
\vspace{-0.2cm}
\end{table}

\subsection{Ablation Study}
We performed ablation experiments on the MultiWOZ dataset, the ablation results are shown in Table \ref{tab: ablation}. \texttt{w/o pre\_training} means directly fine-tuning T5 on the downstream task without TOD pre-training. The results show that the proposed pre-training method brings 4 points of improvements for the MultiWOZ dataset. \texttt{w/o $TPLD$} means the model is trained with the traditional multi-task learning method, which learns all pre-training tasks simultaneously. The pre-training loss is defined in equation (\ref{multi-task}). The \texttt{Combined score} reduced from 102.2 to 98.8 after removing the task-progressive training framework, which indicates that the proposed TPLD multi-stage pre-training framework is crucial for dialog modeling. It is also more difficult for multi-task learning method to optimize the parameters compared to TPLD multi-stage method. 
\begin{equation}
\begin{split}
    \mathcal{L}_{\textit{multi-task}} &= \mathcal{L}_{gen} + \alpha \mathcal{L}_{gpc} + \beta \mathcal{L}_{acl} \\
    \mathcal{L}_{gen} &= \mathcal{L}_{gen\_b} + \mathcal{L}_{gen\_a} + \mathcal{L}_{gen\_r} \\
\end{split}
\label{multi-task}
\end{equation}

For the ablation of the policy-aware pre-training tasks, the combined score decreases by 1.6, 2.3, and 2.6 points after removing $\mathcal{L}_{acl}$, $\mathcal{L}_{session}$, and $\mathcal{L}_{gpc}$, respectively. The model performance further decreases when removing both tasks. The results demonstrate that the two proposed policy-aware pre-training tasks can help the model learn better dialog policy to complete a dialog successfully.

\begin{table}[htb]
    \centering
    \caption{Ablation results on MultiWOZ.}
    \label{tab: ablation}
    \scalebox{0.8}{
    \begin{tabular}{lcccc}
    \toprule
    Model & 
    \texttt{Inform} & \texttt{Success} & \texttt{BLEU} & \texttt{Comb}\\ 
    \midrule
    ours & 89.5 & 77.6 & 18.7 & 102.2 \\
    \ \ \ w/o pre\_training & 86.6 & 72.3 & 18.5 & 98.0 \\
    \ \ \ w/o $TPLD$ & 86.8 & 73.9 & 18.4 & 98.8 \\  
    \midrule
    \ \ \ w/o $\mathcal{L}_{acl}$ & 87.7 & 75.9 & 18.8 & 100.6 \\  
    \ \ \ w/o $\mathcal{L}_{session}$ & 87.1 & 74.8 & 19.0 & 99.9 \\ 
    \ \ \ w/o $\mathcal{L}_{gpc}$ & 87.4 & 74.6 & 18.6 & 99.6 \\
    \ \ \ w/o $\mathcal{L}_{acl} -\mathcal{L}_{gpc}$ & 86.2 & 74.6 & 19.1 & 99.5 \\
    \bottomrule
    \end{tabular}
    }
\vspace{-0.4cm}
\end{table}

\subsection{Loss Decaying Coefficient Analysis}
\begin{figure}[htb] 
    \centering 
    \includegraphics[width=0.5\textwidth]{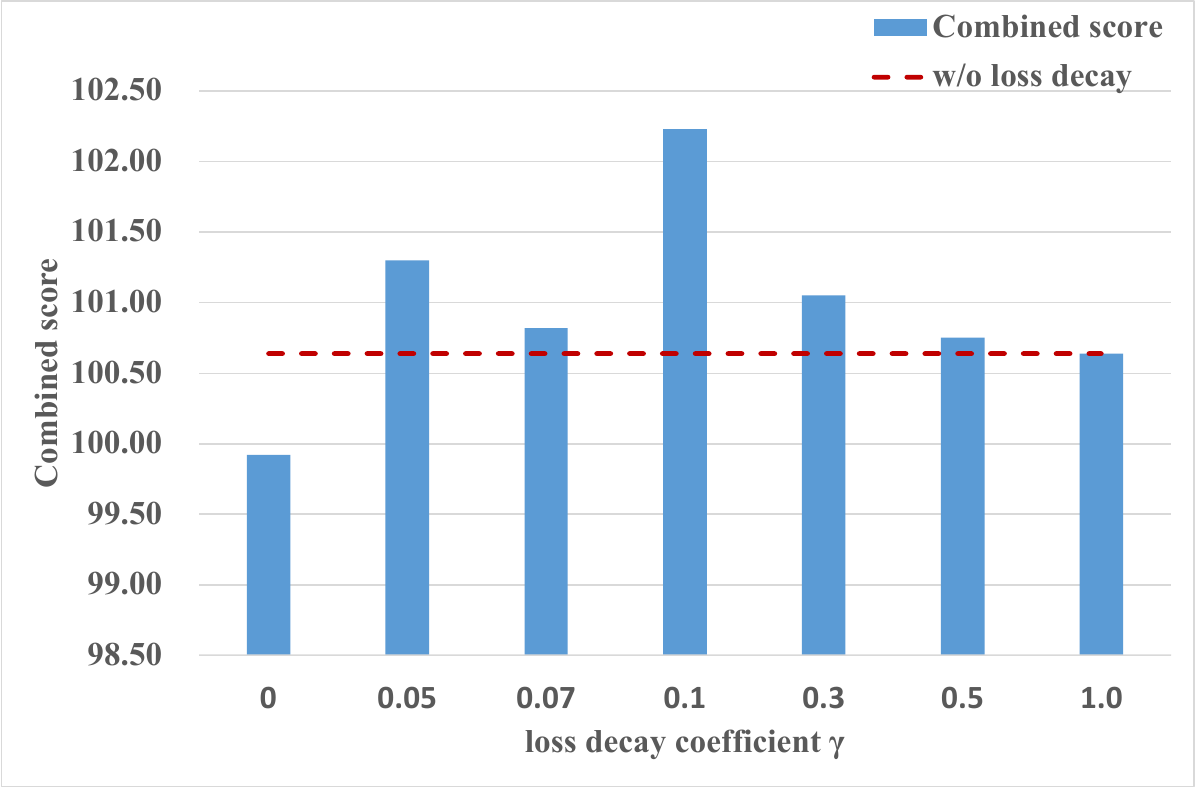} 
    \caption{The analysis of the combined score of different loss decay coefficients.}
    \label{loss decay}
\vspace{-0,3cm}
\end{figure}

Figure \ref{loss decay} shows the effect of different decaying coefficients $\gamma$, where $\gamma$ ranges from 0 to 1.
The model has the worst and the second worst performance at $\gamma = 0$ and $\gamma = 1$. The model achieves the best performance at $\gamma = 0.1$. It demonstrates that the proposed task-progressive with proper loss decaying multi-stage pre-training framework is effective for learning heterogeneous TOD tasks.

\subsection{Case Study}

\subsubsection{Turn-level vs. Session-level.} 
\begin{figure}[htb] 
\centering 
\includegraphics[width=.9 \textwidth]{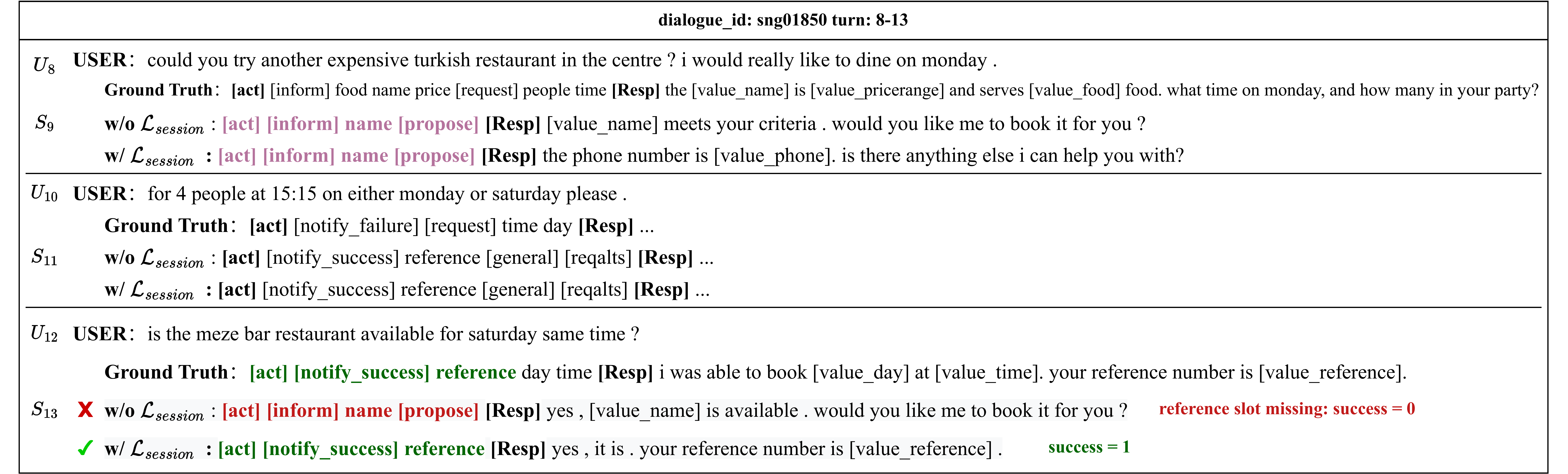} 
\caption{The 8th turn to 13th turn from the dialog session SNG01850 in the test set.}
\label{case study}
\vspace{-0.5cm}
\end{figure}

Figure \ref{case study} shows several output cases of the model with or without $\mathcal{L}_{session}$. The model with $\mathcal{L}_{session}$ avoids generating repetitive system acts to complete the user requests in shorter dialog turns. In the dialog, the user wants to reserve a restaurant. At turn $S_{9}$, the model with $\mathcal{L}_{session}$ or without $\mathcal{L}_{session}$ both \texttt{[propose]} to make a reservation for the user. At turn $U_{12}$, the user modifies the slot value of the restaurant name. The model with $\mathcal{L}_{session}$ knows that the reservation request has been made to the user at turn $S_{9}$, and the user agrees to make the reservation. Therefore, the reservation would be made directly at turn $S_{13}$ and provide \texttt{reference}. However, the model without $\mathcal{L}_{session}$ does not know what system act has been generated before. Therefore, the model will repeat the same system act as turn $S_{9}$, which will miss the chance to provide \texttt{reference} to the user, and the dialog will fail.

\vspace{-0.3cm}

\section{Conclusion}
This paper proposes a novel TPLD multi-stage pre-training framework for training TOD PCMs. The TPLD framework progressively trains the DST, POL, and NLG tasks through three successive stages. We also design two policy-aware pre-training tasks as POL tasks to model the multi-turn dialog policy sequence and policy similarity between samples during pre-training, respectively. Experiments show that our model achieves new state-of-the-art results on MultiWOZ and In-Car end-to-end dialog modeling benchmarks compared with other strong PCMs. We hope that TPLD multi-stage pre-training framework and policy-aware pre-training tasks can push forward the research in the task-oriented dialog pre-training area as well as the design for Large Language Models(LLMs) for TOD.

\section*{Acknowledgements}
We are grateful to the anonymous reviewers for their insightful comments and suggestions.

%
%
%
%

\end{document}